\documentclass[10pt,twocolumn,letterpaper]{article}

\usepackage[pagenumbers]{cvpr} 

\usepackage{richeek}
\renewcommand \paragraph[1] {\medskip \noindent \textbf{#1}}

\DeclareSIUnit{\events}{\text{events}}
\def \disp {\text{d}}

\graphicspath{{./images/}}

\definecolor{cvprblue}{rgb}{0.21,0.49,0.74}
\usepackage[pagebackref,breaklinks,colorlinks,allcolors=cvprblue]{hyperref}

\def\paperID{*****} 
\def\confName{CVPR}
\def\confYear{2026}

\title{Neurosim: A Fast Simulator for Neuromorphic Robot Perception}

\author{Richeek Das \quad\quad Pratik Chaudhari \\
GRASP Laboratory, University of Pennsylvania \\
{\tt\small \{richeek, pratikac\}@seas.upenn.edu}
}

\begin{document}
\maketitle
\begin{abstract}
Neurosim is a fast, real-time, high-performance library for simulating sensors such as dynamic vision sensors, RGB cameras, depth sensors, optical flow, and inertial sensors. It can simulate agile dynamics of multi-rotor vehicles in complex and dynamic environments. Neurosim achieves frame rates as high as 2700 FPS on a desktop GPU, with $\sim$10$\times$ faster event simulation than existing methods. Neurosim integrates with a ZeroMQ-based communication library called Cortex to facilitate seamless integration with machine learning and robotics workflows. Cortex provides a high-throughput, low-latency message-passing system for Python and C++ applications, with native support for NumPy arrays and PyTorch tensors. This paper discusses the design philosophy behind Neurosim and Cortex. It demonstrates how they can be used to (i) train neuromorphic perception and control algorithms, e.g., using self-supervised and reinforcement learning on time-synchronized multi-modal data, and (ii) test real-time implementations of these algorithms in closed-loop. \footnote{Neurosim and Cortex are available at \href{https://github.com/grasp-lyrl/neurosim}{https://github.com/grasp-lyrl/neurosim}.}
\end{abstract}    
\section{Introduction}
\label{sec:intro}

\begin{figure}[t!]
    \centering
    \includegraphics[width=\linewidth]{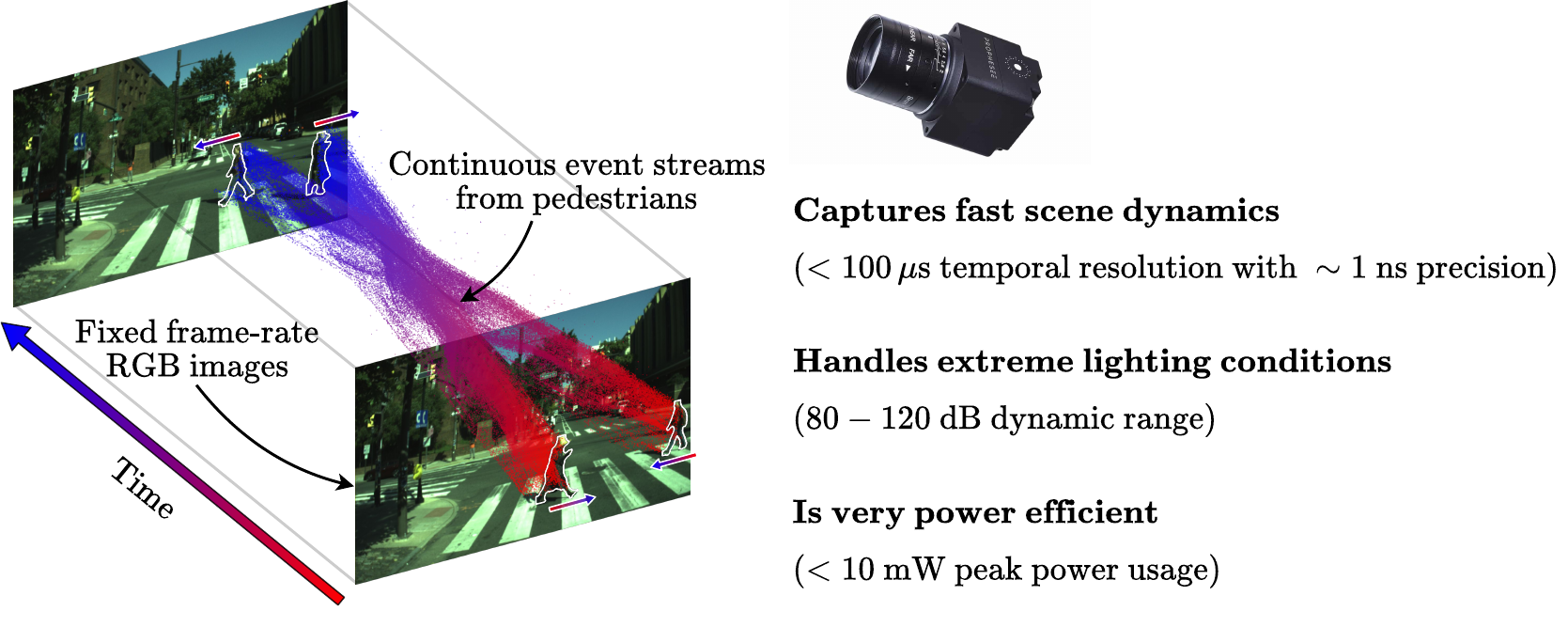}
    \caption{\textbf{Event cameras capture the scene without temporal aliasing, with high dynamic range, and consume very little power.} We illustrate the temporal non-aliasing property of event cameras in the left image. Unlike standard RGB cameras, event cameras are asynchronous sensors that respond to scene intensity changes and produce a continuous stream of information at high temporal and spatial precision.}
    \label{fig:events}
    \vspace{-1\baselineskip}
\end{figure}

Beginning with the work of Carver Mead and Misha Mahowald in the late 1980s, researchers in neuromorphic engineering have sought to emulate the retina and the rest of the brain in silicon. Event-based cameras are the latest hardware that this research has produced. They emulate rods and cones in the retina and record changes in the intensity of light (increases or decreases beyond a threshold) at each pixel. Event cameras, as illustrated in \cref{fig:events}, have a high temporal resolution ($<$ 100 \unit{\us}), high dynamic range (80-120 \unit{\dB}) and operate asynchronously with high temporal precision ($\sim$1 \unit{\ns}). These properties make them ideal for fast, robust perception in difficult lighting, high speeds, and power-constrained settings.

Data from an event camera consists of a stream of 4-tuples $(t_i, u_i, p_i)$ for $i \in \naturals$ where $t_i \in \integers_+$ denotes the timestamp (typically, in microseconds) of an ``event'', $u_i \in \integers^2$ are the pixel coordinates where the event occurred and the polarity $p_i \in \{-1, 1\}$ indicates whether the intensity at pixel $u_i$ increased or decreased. So the first challenge in developing algorithms for event data comes from the fact that we need to rethink computer vision from the ground up---unlike a standard RGB camera, an event camera does not register an image. Even annotating event data for supervised learning on downstream tasks is challenging, due to its unusual nature and because it is quite noisy. It is tempting to use cross-modal supervision, e.g., either by transferring annotations from RGB and LiDAR data or by using pseudo-labels from systems trained on these other modalities. This is a viable approach, but provided appropriate care is taken to synchronize timestamps and obtain accurate extrinsic calibration. The largest event camera dataset containing multi-sensor data, M3ED, has less than four hours of event data \cite{Chaney_2023_CVPR}.

Simulators enable us to work around these issues. Our desiderata for a simulator are as follows.
\begin{enumerate}
\item \textbf{Efficient, high-fidelity event simulation.}
Events sense only the changes in the intensity of light, so we should expect that the sim-to-real gap for systems trained on simulated event data is smaller than that trained on simulated RGB data. RGB simulators need to capture the rich color information, lighting effects (shadows, reflections, etc.), and motion blur to be effective in the real world. Works like \cite{klenk2023devo} have obtained remarkable results by training entirely on simulated event datasets.
We need to render the scene at very high frame rates (a good thumb-rule is 1000+ FPS) for high-fidelity event simulation. Ideally, event data should be created on-the-fly without storing the corresponding rendered RGB images.%
\footnote{Existing simulators like TartanAir \cite{tartanair2020iros} work around this problem by generating high frame rate (1000 FPS) RGB data, saving them to disk, and then post-processing them to generate event data using \cite{esim}---a prohibitively slow process that consumes roughly 2-3 terabytes of storage for an hour of high-definition (HD) simulation.
Alternatives like CARLA \cite{Dosovitskiy17} and AirSim \cite{airsim2017fsr} suffer from temporal aliasing. They rely on low frame rate ($\sim$30 FPS) RGB rendering to approximate high frequency events, and as a consequence have significant artifacts in the generated event data.}

\item \textbf{High-throughput multi-modal data from sensors for RGB images, depth, optical flow, and inertial motion, along with ground-truth from the scene and intrinsic and extrinsic calibration.}
For example, a typical robot receives LiDAR data at $\sim$10 \unit{\Hz} and in total about a half a billion points per second, event data from a monocular high-definition camera at $\sim$50 M events per second, an HD RGB camera provides about $\sim$1 M pixels at $\sim$100 \unit{\Hz}, while an inertial measurement unit can register angular velocities, linear accelerations and magnetometer orientations upwards of $\sim$500 \unit{\Hz}. Sensors typically operate using their individual clocks and are synchronized externally. The quality of synchronization depends on the kind of hardware used for this purpose, and also upon the estimated motion of the robot. A simulator should simulate these realities of real-world robots that will affect downstream algorithms.
And do so extremely quickly in a fraction of real-time for simulation to be a viable alternative to data curation.

\item \textbf{Feeding multi-modal robot perception data to deep learning training pipelines.}
The sensors above would result in about 1 \unit{\giga\byte\per\s}. It is important to design mechanisms to feed such low-latency, high-throughput data to deep learning training pipelines in Python-native formats without having to store the data on a hard disk, or excessive buffering in the data loaders, or artificially slowing down the simulator.

\item \textbf{Ability to run closed-loop perception and control experiments at the extremes of the performance envelope of the hardware.}
Event cameras lead to interesting perception-control loops for agile robots. While RGB images would be hopelessly blurred when a quadrotor undergoes a flip (angular velocity of $\sim$700 \unit{\degree\per\s}), an event camera would see key-points without temporal aliasing. It is difficult to set up experiments that push the hardware platform to the limits of its agility, e.g., flips, avoiding moving obstacles, or a humanoid falling, it risks damage to the platform and the researcher.
\end{enumerate}


\section{The Design of Neurosim}
\label{sec:neurosim}






\begin{figure*}
    \centering
    \includegraphics[width=0.9\linewidth]{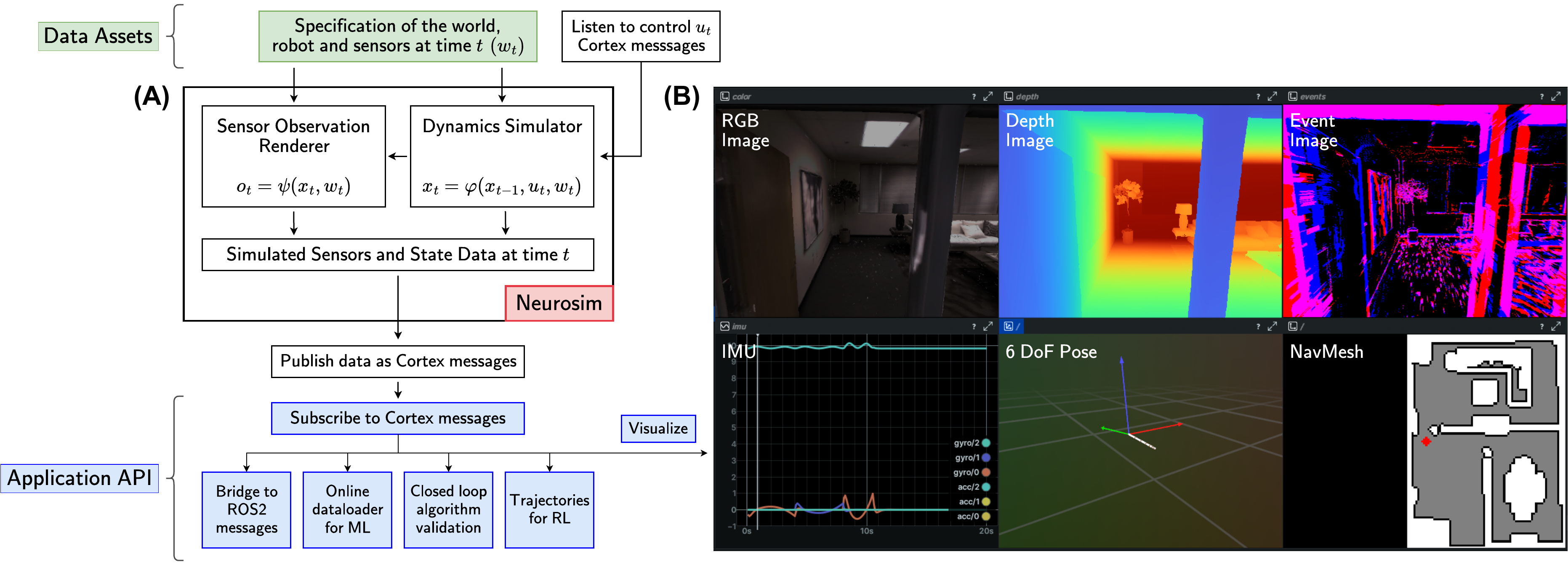}
    \caption{\textbf{Overview of the design of Neurosim.} Neurosim is designed to be modular, high-performance, and easy to use for a variety of applications in embodied perception on multirotors.
    As shown in \textbf{(A)}, it consists of four main components:
    (1) A data stack in the form of 3D scene assets and the specifications for sensors and multirotors,
    (2) A real-time rendering engine for high-fidelity vision sensor simulation,
    (3) A fast and accurate multirotor dynamics model for aerodynamics and physics, and
    (4) A communication interface to interact with the simulator -- receive sensor data and send control commands.
    In \textbf{(B)}, we illustrate how we can use the communication interface to connect Neurosim to a visualizer -- here Rerun \cite{RerunSDK} -- and draw the simulated data in real-time. It displays the rendered RGB image, depth map, IMU readings, 6-DoF pose, navigation mesh, and the event stream. For visualization, events are accumulated over short time windows (20 \unit{\ms}) and colored by polarity.
    }
    \label{fig:neurosim_architecture}
    \vspace{-1\baselineskip}
\end{figure*}


Neurosim is an asynchronous, modular, and high-performance simulator designed with a focus on neuromorphic perception on multirotors.
Its design philosophy hinges on four key principles:
(1) modularity -- each component (rendering, dynamics, controllers, communication) operates independently and can be extended, replaced, and fault-isolated;
(2) low-latency -- the visual and dynamics simulators are chosen to be fast. The full simulation stack (with an event camera) achieves $\sim$2700 FPS on a single Nvidia 4090 GPU;
(2) real-time streaming -- data flows directly to training pipelines without the need for large intermediate disk storage. Reproducibility is ensured through randomization seeds rather than traditional static datasets;
(4) Pythonic accessibility -- a simple API that is Python and deep learning friendly. As we will show in \cref{sec:applications}, these design choices make Neurosim easy to use for a variety of applications in embodied perception.
In this section, we discuss the key components of Neurosim as illustrated in \cref{fig:neurosim_architecture}.

\paragraph{Neurosim simulates event cameras at multi-kilohertz rates.}
We choose Habitat-Sim \cite{habitat19iccv} to render the scene because of its exceptional rendering speed and flexibility. It supports diverse 3D asset formats, is compatible with multiple open-source 3D scene datasets, and provides built-in rigid-body physics integration through PyBullet \cite{coumans2016pybullet}. Habitat-Sim is equipped with a variety of sensors and camera models, including RGB-D, semantic, and egomotion cameras -- each configurable with independent intrinsics and sensor parameters.
Habitat-Sim can render a typical indoor scene with multiple RGB-D sensors at VGA (640 $\times$ 480) resolutions at $\sim$3000 FPS on a single GPU.%
\footnote{Our modular design allows the visual renderer in Neurosim to support diverse 3D assets and sensors. For Habitat-Sim, this includes Matterport3D, Gibson, Replica, MP3D+Gibson, etc., and sensors such as RGB, Depth, Semantic, Egomotion, etc.}

Neurosim uses the rendered intensity images to simulate an event camera by keeping track of a state for each pixel that records the intensity corresponding to the last triggered event at that pixel. The high frame rate in Habitat-Sim reduces the likelihood of large intensity changes between consecutive frames that could lead to missed events.
Real-world event sensors exhibit non-idealities such as noise, bandwidth limitations, and sensor saturation effects. Neurosim simulates these artifacts, although they can be disabled by the user if so desired.
The resultant rendering speed, more than 2700 FPS, corresponds to a temporal resolution of $\sim$0.37 \unit{\ms} for simulated events. Most event-based algorithms working on real data utilize event rates far lower than our simulated resolution
\cite{das2025fastfeaturefield}, \cite{evsurvey}.

Current CPU implementations of the above contrast-threshold-based event generation model in the literature \cite{Dosovitskiy17,airsim2017fsr} are exceedingly slow. GPU implementations such as \cite{esim} are faster, but they still require multiple CUDA-kernel launches and do not effectively utilize the shared memory and warp parallelism features of modern GPUs.
Neurosim uses a CUDA kernel to compute events across pixels in parallel. It then aggregates events efficiently at warp level to avoid atomic contention. Each thread first processes a single pixel and determines whether an event is triggered. Threads within a warp then cooperatively compute a bitmask of active (event-generating) threads, and count the number of events in that warp. GPU SRAM (shared memory, which is roughly 20$\times$ faster than global memory) is used to store intermediate results and perform warp-level reductions.
A single atomic operation reserves a contiguous block in the global event buffers for all events in that warp. The base index of this block is broadcast to all threads in the warp. Each active thread then computes its own offset within the warp by counting preceding set bits in the warp mask, and writes its event (pixel coordinates, timestamp, and polarity) into the pre-allocated segment of the output arrays. This warp-synchronous design minimizes the number of global atomics from one per event to only about one per warp -- roughly a 32$\times$ (warp size) reduction. When aggregated across all resident warps on a streaming multiprocessor (e.g., 64 warps per SM on recent Nvidia architectures), the peak number of concurrent event writes can scale to thousands of threads per SM while still using only tens of atomics. This is important because each rendering step produces a variable number of events. Naive pre-allocation requires two kernel launches, and per-event atomics are prohibitively expensive.

\begin{figure*}
    \centering
    \includegraphics[width=0.9\textwidth]{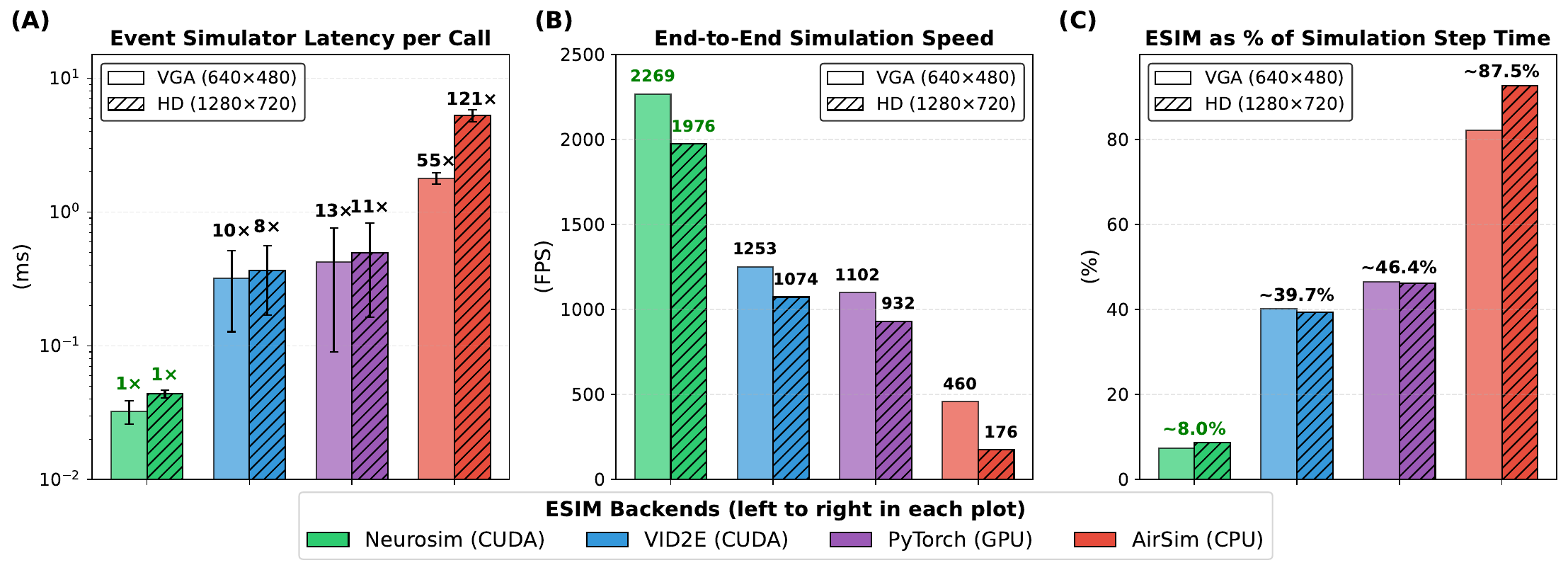}
    \caption{\textbf{Neurosim event simulator backend is optimized to prevent bottlenecking of the full simulation pipeline.}
    We benchmark Neurosim's event camera simulator against existing simulators: a CUDA-based simulator \cite{esim}, a GPU-based PyTorch implementation, and a CPU-based implementation \cite{airsim2017fsr}.
    \textbf{A.} We measure the average latency to simulate events from a single intensity image update -- the time taken to process a new intensity frame, trigger events based on the contrast threshold model, update the internal pixel states, and output the event stream.
    Neurosim achieves over 31 kHz for VGA and 23 kHz for HD frame sizes. It is roughly 8--13 $\times$ faster than other GPU-based implementations, and 55--121 $\times$ faster than CPU-based implementations.
    \textbf{B.} Neurosim using the custom event simulator backend achieves $\sim$2300 FPS for a full simulation step (including rendering, dynamics, and event simulation) with VGA vision sensors. Other event simulation backends bottleneck the full simulation pipeline, achieving only 200--1200 FPS.
    \textbf{C.} Neurosim's event simulator only takes up $\sim$ 8 \% of the total simulation step time compared to 40--90 \% for other event simulators.
    This high performance is attributed to the warp-synchronous CUDA kernel design that minimizes atomic operations during event aggregation and allows our event generation model to run in a single kernel launch.
    }
    \label{fig:neurosim_esim_benchmark}
    \vspace{-1\baselineskip}
\end{figure*}

Our design significantly increases throughput and sustains event generation from multi-kilohertz intensity images on modern GPUs. Neurosim can process a new intensity frame, trigger events based on the contrast threshold model, update the internal pixel states, and output the event stream at over 31 kHz for VGA and 23 kHz for HD frames.
We benchmark our implementation against existing event simulators in \cref{fig:neurosim_esim_benchmark}.

\paragraph{Running closed-loop perception and control experiments at the extremes of the performance envelope of the hardware}
Neurosim, in its first iteration, includes support for a fast and accurate model of multirotor dynamics and inertial sensors named RotorPy \cite{folk2023rotorpy}. Control commands (rotor RPMs, or thrust and angular velocities in body frame) can be provided to change the vehicle pose. We include parameters like aerodynamic drag specifications and multirotor configurations like mass, inertia, rotor positions and motor response characteristics for accurate simulation. Neurosim also supports swapping out the dynamics simulator for different libraries, custom implementations, or other platforms.
RotorPy can simulate the dynamics at more than 20 \unit{\kilo\Hz} on a single CPU core. It includes procedures for motion planning for multirotor platforms using differential flatness-based trajectory generation. Dynamics and sensor simulation in Neurosim are coupled tightly with very little latency to ensure that sensor simulation faithfully captures the vehicle pose. \cref{sec:applications} shows an example of how this can be used to sample agile trajectories in congested indoor environments to simulate sensory data that would be much too risky to obtain using real hardware.

The Neurosim architecture mirrors a typical autonomy stack that interfaces with a hardware platform. Different processes consume sensor observations, compute upon them using their respective perception, planning, or control pipelines, and publish control commands. The simulator ingests these commands just like a real multirotor would. There are two modes of operation. In the ``closed-loop mode'', Neurosim runs in lockstep with other processes that subscribe to sensor streams, perform computation, and publish control commands that immediately affect the simulated vehicle. Example applications of such synchronous execution include evaluating event-based perception and control algorithms or reinforcement learning event-based control policies. In the ``streaming mode'', Neurosim behaves as if a platform were transmitting real-time data for online training. \cref{sec:applications} demonstrates an example application of asynchronous simulation. The simulator publishes observations that are used directly for online training of multi-modal foundation models without any intermediate disk storage.  \cref{sec:cortex} discusses how high-frequency and high-throughput perception and control data (e.g., Python data structures, NumPy, and PyTorch arrays) for both synchronous and asynchronous operation are communicated using a publish-subscribe framework called Cortex.

\section{The Cortex Communication Interface}
\label{sec:cortex}


Cortex is a high-performance communication library for robotics and machine learning applications that is designed to work together with Neurosim. It connects the simulator to different kinds of downstream data consumers and connects the controllers back to the simulator. Cortex is designed to be a lightweight, low-latency and compatible alternative to middleware like ROS \cite{ros2}, while providing a more robotics-friendly abstraction than raw messaging libraries like ZeroMQ \cite{zmq} or plain Unix sockets.

The core philosophy behind Cortex is as follows. First, modern CPUs and memory subsystems have evolved dramatically in recent years. For example, consumer-grade desktop AMD processors have more than 128 \unit{\mega\byte} L3 cache memory, clock speeds upwards of 4 \unit{\giga\Hz}, and more than 16 cores. This makes low-latency, zero-copy inter-process communication (IPC) extremely efficient, without having to perform frequent memory accesses. Second, there are a number of performant low-level C/C++ based libraries that utilize these hardware advancements, e.g., ZeroMQ or NanoMsg. They also come with efficient Python bindings. Python 3.14 introduces free-threading support by removing the Global Interpreter Lock (GIL) that has historically hindered multi-threaded parallelism. So Python interpretation is no longer the bottleneck for low-latency, high-throughput inter-process communication. Coupled with new techniques like just-in-time (JIT) compilation, Python is becoming viable for high-performance scripting.

\paragraph{Feeding high-throughput multi-modal robot sensory data to deep learning training pipelines.}
As its transport layer, Cortex uses ZeroMQ \cite{zmq}, a high-performance asynchronous messaging library. This is a robust and efficient communication substrate, supporting various patterns (e.g., publish-subscribe, request-reply) and protocols (e.g., TCP, IPC, PGM). A key feature of Cortex is dynamic node discovery. A lightweight discovery daemon runs in the background and maintains a registry of all active nodes and their advertised topics. Nodes register themselves with the daemon upon startup and query it to find other nodes and topics they wish to communicate with.
\begin{figure}
    \centering
    \includegraphics[width=\linewidth]{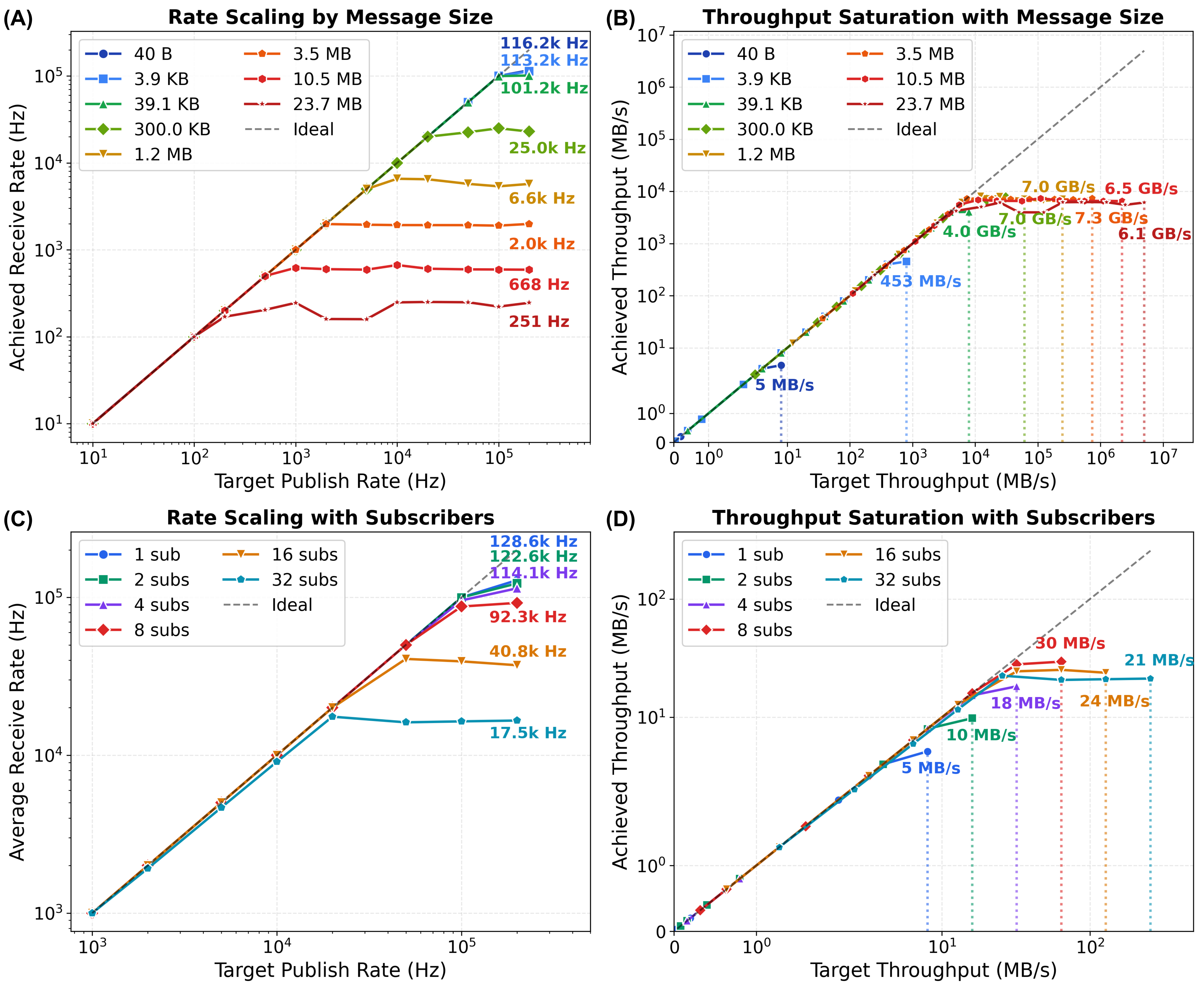}
    \caption{\textbf{Cortex provides high-throughput, low-latency, and scalable communication in Python.}
    We benchmark different aspects of Cortex's performance on a modern desktop CPU (AMD Ryzen 9 7950X).
    \textbf{A.} We use NumPy float (32 \unit{\bit}) arrays of different sizes ranging from 10 elements (40 \unit{\byte}) to 1080$\times$1920 RGB images (23.7 \unit{\mega\byte}) as our message payloads. We measure the max achieved successful publish-subscribe rate (in \unit{\hertz}) for each payload size over a 15 \unit{\second} interval. Cortex achieves 100+ \unit{\kilo\hertz} for small ($<$ 39 \unit{\kilo\byte}) messages and maintains sufficiently high rates (250 \unit{\hertz}) for large 1080p RGB images.
    \textbf{B.} We measure the achieved throughput (in \unit{\mega\byte\per\second}) for the same payload sizes. Cortex achieves up to 7 \unit{\giga\byte\per\second} throughput for large messages, saturating the underlying hardware limits. For smaller messages, the throughput is limited by per-message overheads -- the messaging rate caps at $\sim$100 \unit{\kilo\hertz}.
    \textbf{C.} We test the scalability of Cortex by varying the number of subscribers from 1 to 32, each receiving 40 \unit{\byte} messages from a single publisher. The achieved message rate per subscriber remains constant at around 100 \unit{\kilo\hertz} as the number of subscribers increases up to 8, after which it degrades due to CPU contention from many busy-waiting subscribers on the same machine.
    \textbf{D.} We measure the achieved throughput (in \unit{\mega\byte\per\second}) for the same setup as in C. As deduced from the message rates, the throughput scales roughly linearly with the number of subscribers until CPU contention effects kick in beyond 8 subscribers.
    }
    \label{fig:cortex_benchmark}
    \vspace{-1.25\baselineskip}
\end{figure}
This allows for dynamic and flexible network topologies without a centralized broker or hardcoded addresses. Communication between processes remains peer-to-peer in nature; this is critical to minimizing latency and single points of failure. Processes can use different transports depending on their use cases, and multiple subscribers can listen to the same topic seamlessly.
We benchmarked Cortex on a modern desktop CPU (AMD Ryzen 9 7950X) in \cref{fig:cortex_benchmark} to find that it achieves message rates, throughput, and scalability that is suitable for high-bandwidth robotics and machine learning applications.

Cortex is designed with native support for libraries like NumPy and PyTorch. This allows efficient serialization and deserialization of complex data structures such as multi-dimensional arrays and tensors. Cortex employs zero-copy techniques for these messages with exposed memory buffers, as long as either the producer or consumer ensures memory safety. Inspired by the philosophy of LCM \cite{lcm2010}, Cortex provides a simple messaging API with strong type checking. Users define custom message types using Python dataclasses with NumPy and PyTorch support, rather than being restricted to slower alternatives like Python lists. The library handles the details of serialization, deserialization, and transport. For type safety, Cortex computes a unique 64-bit hash for each message type based on its structure, fields, and data types. This hash is included in the message header, allowing receivers to verify that they are receiving the expected message type, catch mismatches early, and dynamically load/cache deserialization logic for new message types at runtime. This ensures strong type safety and helps prevent runtime errors due to incompatible message formats.

A majority of robotic autonomy code is implemented within the ROS ecosystem today. Cortex provides a performant bidirectional bridge to ROS 2. The bridge allows researchers and developers to use Cortex as a lightweight communication layer for high-bandwidth machine learning data streams while still enjoying the extensive support of the ROS 2 ecosystem.
In the context of Neurosim, this bridge allows users to use ROS 2 workflows, say for controllers and state estimators, seamlessly, while benefiting from Cortex-based communication in Neurosim behind the scenes. For example, using this Cortex-based interface, one can use Rerun \cite{RerunSDK} while simultaneously using RViz in ROS 2 \cite{ros2} for visualizing some user-specific code artifacts-- both receiving data and synchronized against the same underlying Neurosim instance.

Cortex also facilitates efficient data loading for downstream applications by removing disk I/O bottlenecks.
Neurosim behaves like a robot platform that sends real-time rendered sensory data over Cortex -- broadcasting multi-modal sensor streams like RGB, depth, events, IMU, and state estimates.
Potential consumers of this data stream are deep learning training pipelines that ingest the data for training.
Data subscribers operate in separate processes, utilizing zero-copy deserialization to receive data using Cortex and push it into multiprocessing queues.
Online data loaders consume these queues and assemble the data into pre-allocated PyTorch tensors. This architecture handles modality-specific batching (such as aggregating variable-length event streams), prefetching of data batches, and multi-GPU distributed training.
Cortex-based online data loading is significantly more scalable than disk-based loading. It allows a single simulation node to saturate the ingestion bandwidth of multiple training GPUs and removes the logistical burden of managing terabytes of static simulated data.

\section{Applications}
\label{sec:applications}

The combination of Neurosim and Cortex enables a range of applications that were previously impractical due to limitations of existing tools. We next present three kinds of experiments that demonstrate the capabilities of our framework.

\begin{figure*}
    \centering
    \includegraphics[width=0.79\linewidth]{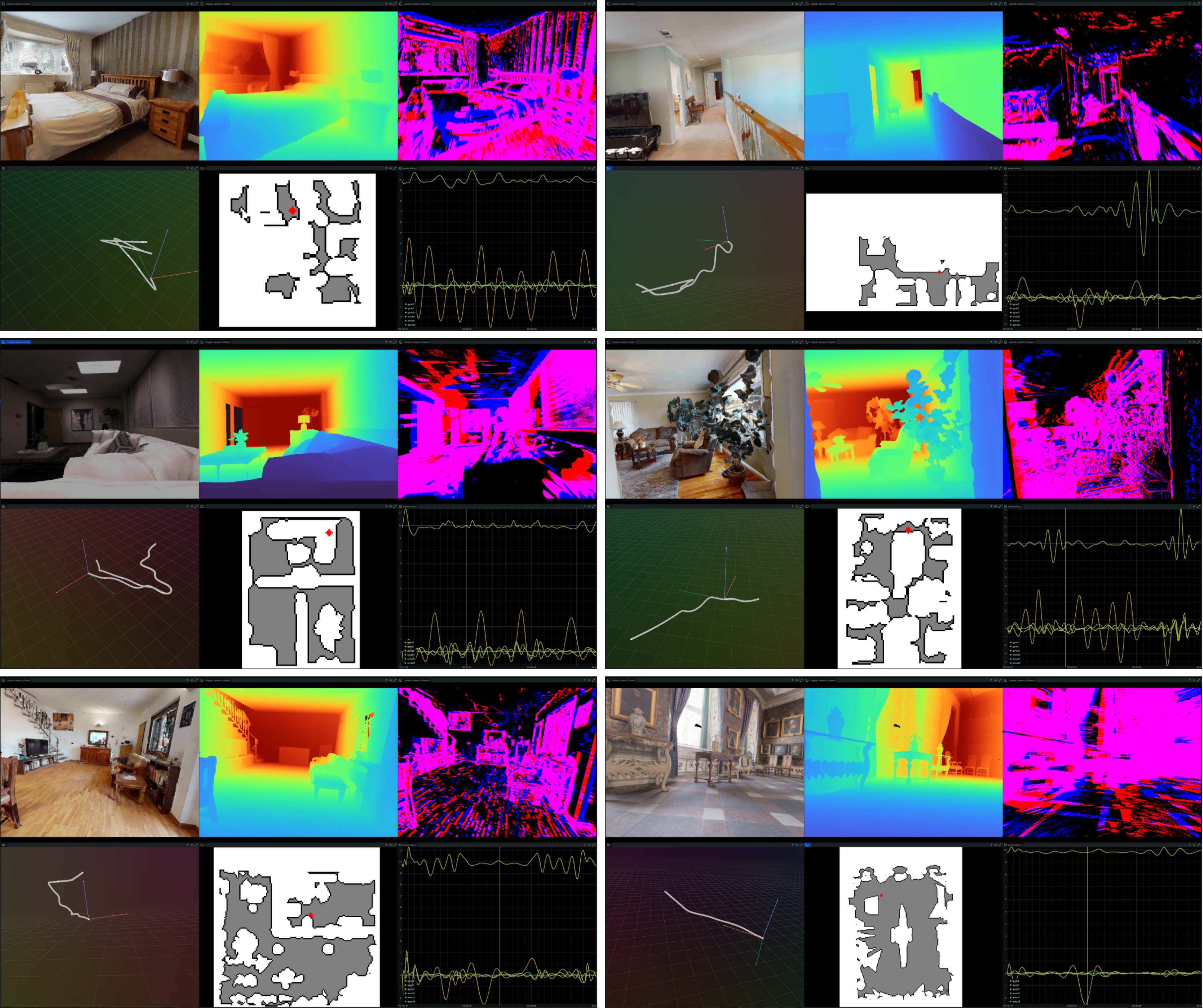}
    \caption{
    \textbf{Examples of a quadrotor tracking randomly generated MinSnap trajectories in a variety of indoor scenes.}
    In each example, from left to right, we plot the color image, depth image (near to far is colored as dark blue to red), and events aggregated with polarity over 20 \unit{\milli\second}, on the top. 3D trajectory with current pose, navigation mesh with current position, and IMU readings are plotted at the bottom. The navigation mesh marks free space and non-reachable locations at the current flying height with gray and white, respectively. These scenes demonstrate fast flights across multiple floors, corridors, and interconnected rooms, with angular and linear velocities often exceeding 4 \unit{\radian\per\second} and 2 \unit{\meter\per\second}, respectively.
    }
    \label{fig:neurosim_examples}
    \vspace{-1\baselineskip}
\end{figure*}

\paragraph{Real-time, closed-loop control.}
Neurosim facilitates the development of end-to-end control algorithms in a real-time and closed-loop setting using asynchronous simulator and controller processes connected via Cortex.
As illustrated in \cref{fig:neurosim_architecture}, the simulator broadcasts multi-modal sensor data, which are consumed by a controller to generate actuation commands, such as individual rotor speeds. In this mode of operation, the latency of control algorithms plays an important role in closed-loop performance.
Slow controllers and communication bottlenecks can be simulated, and the effect of such limitations can be studied precisely. By default, the simulator performs a zeroth-order hold on the most recent control command if new commands are not received in time.
This is useful to evaluate algorithms on different hardware platforms, e.g., a simulator node can run on a powerful desktop GPU, while the controller can be tested on actual robot hardware such as NVIDIA Jetson Orin.
This setup is particularly valuable for debugging and testing reactive algorithms, such as event-based collision avoidance or high-speed visual servoing, where controllers must respond to environmental changes with millisecond precision.
In \cref{fig:neurosim_applications} (A), we analyze the control-to-simulation latency and achievable sensor rates for a closed-loop controller receiving RGB images from Neurosim at different controller rates up to 5 \unit{\kilo\Hz}.
In this analysis, both the controller and simulator nodes are running on the same machine and communicate over Cortex using IPC. These results indicate that Cortex can sustain low-latency communication ($<$ 0.7 \unit{\milli\second}) and high sensor rates ($\sim$ 2.3 \unit{\kilo\hertz}), making it suitable for real-time closed-loop control applications. In effect, this analysis shows that Neurosim supports evaluation and development of real-time closed-loop control at more than 1500 \unit{\Hz}.

The modular design of Neurosim, enabled by Cortex, allows for flexibility in the control stack. The simulator is agnostic to the controller's implementation. One can prototype in Python, deploy high-performance C++ implementations, or interface with existing ROS 2 navigation stacks. This extensibility can support a wide range of use cases, from benchmarking end-to-end learning-based policies to performing safety analysis and regression testing of control algorithms.

\paragraph{Online training of multi-modal (including event-based) perception models.}
We demonstrate the utility of Neurosim by training an event-based monocular depth estimation network using the synchronized data generated in simulation. This network uses Fast Feature Field ($\text{F}^3$) as the input representation that was developed by  Das et al. \cite{das2025fastfeaturefield}. In short, given a set $e^- = \{(s, u): e(s, u) \in \{+1, -1\} \text{ for } s \in [t-\D t, t), u \in \Omega \}$ that consists of events of positive or negative polarity within a time horizon $[t-\D t, t)$ and the image plane $\Om \in \integers^2$, the representation of these events $\text{F}^3(t, u) \in \reals^p$ is sufficient to reconstruct future events $e^+$ that correspond to times $s \in [t, t+\D t)$. Das et al. developed a neural architecture to compute $\text{F}^3$ efficiently (at 120 \unit{\Hz} on HD frames and 440 \unit{\Hz} on VGA frames). They showed state-of-the-art performance on downstream tasks ranging from optical flow, stereo depth, monocular depth estimation, and semantic segmentation being performed at 25--75 \unit{\Hz} on HD frames using this representation.

We procedurally generated diverse minimum-snap \cite{minsnap2011} trajectories through complex indoor environments, incorporating a wide variety of yaw profiles and flight speeds. These trajectories are dynamically feasible for the quadrotor platform of choice---accounting for mass, inertia, motor profiles, and aerodynamics---tracked by an SE(3) controller which is provided state estimates at 100+ \unit{\hertz}.
The training pipeline consumes this stream via the online data loader discussed in \cref{sec:cortex}. It receives time-synchronized event streams from a simulated event camera along with ground-truth depth maps.
In addition to the benefits mentioned earlier, this decoupled approach allows us to easily modify the training inputs---scenes, trajectories, sensor parameters---on-the-fly without having to re-generate and store large datasets on the hard disk. Using a visualizer node from Cortex, one can also monitor the simulation in real-time while training is ongoing.
We show some example trajectories and the corresponding sensor data in \cref{fig:neurosim_examples}. These trajectories demonstrate fast flights across multiple floors, corridors, and interconnected rooms, with angular and linear velocities often exceeding 4 \unit{\radian\per\second} and 2 \unit{\meter\per\second}, respectively.

\begin{figure}
    \centering
    \includegraphics[width=\linewidth]{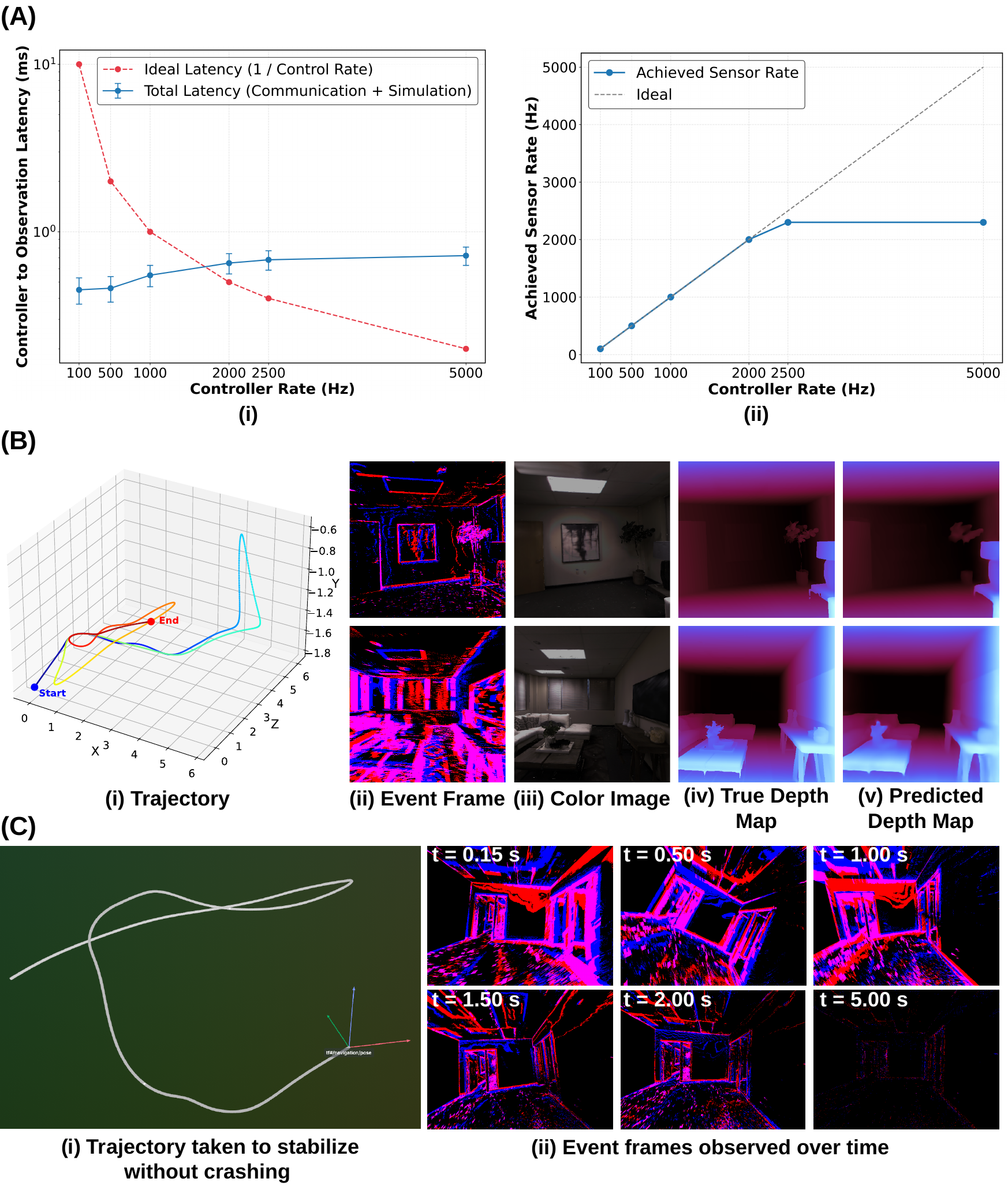}
    \caption{\textbf{Neurosim can be used for closed-loop controller testing, online data generation for learning and on-policy reinforcement learning.}
    \textbf{(A)} shows a latency and frequency analysis of Neurosim in typical indoor flying scenes for closed-loop control. The simulator runs at 10 \unit{\kilo\Hz}, performing dynamics updates. The controller and simulator nodes work separately and communicate asynchronously using Cortex. We vary the controller rate (same as the sensor rendering rate) from 100 \unit{\Hz} to 5 \unit{\kilo\Hz} to measure the effects of communication-simulation latencies and bandwidth saturation. We use a VGA (640 $\times$ 480) RGB camera as our sensor, that is, each sensor observation is $\sim$ 0.9 \unit{\mega\byte} in size.
    \textbf{(i)} shows the average end-to-end latency experienced by the controller -- the time taken from sending a control command to receiving the next sensor observation after the command has been applied. This latency captures the communication overhead as well as the simulator processing time. As the controller and target sensor rate increase, the communication latency increases slightly due to increased IPC bandwidth contention. However, the overall latency remains below 0.7 \unit{\ms} even at 5 \unit{\kilo\Hz} controller rates. We also plot the ideal latency at a particular control rate, that is, the time taken to take control in real-time. Any control rate achieving a controller-to-observation latency less than the ideal latency is guaranteed to receive the latest sensor observations at the control rate.
    \textbf{(ii)} shows the sensor observation rate received by the controller vs the desired sensor rate. The controller is able to receive the VGA color sensor readings at the target rate up to 2.3 \unit{\kilo\Hz}, after which it plateaus. This plateau is the current limit of the simulator's rendering speed on a single Nvidia 4090 GPU, combined with the communication overhead of Cortex. 
    \textbf{(B)} shows qualitative examples of learning event-based monocular depth.
    \textbf{(i)} shows an example MinSnap trajectory sampled in a complex indoor environment, avoiding obstacles. For visualization, the trajectory is
    }
    \label{fig:neurosim_applications}
\end{figure}
\begin{figure}
\ContinuedFloat
\centering
\caption{
    colored based on time, from blue (start) to red (end). \textbf{(B)} \textbf{(ii)} shows 20 \unit{\milli\second} of events plotted onto a frame and colored based on their polarity. The rendered events depend on the underlying motion of the robot -- tracking randomly sampled trajectories while respecting the vehicle dynamics and controllers.
    \textbf{(iii)} and \textbf{(iv)} show the color and depth image captured at the last event timing, respectively.
    \textbf{(v)} shows the depth image predicted from the events plotted in (ii).
    The predictor -- an $\text{F}^3$-based monocular depth model -- is trained online using the data generated from Neurosim.
    \textbf{(C)} shows an example rollout using a policy learnt to stabilize and not crash a quadrotor when thrown with a random initial and angular velocity in a cluttered indoor environment in Neurosim. Here, the quadrotor is initialized with a linear velocity and yaw rate of 2 \unit{\meter\per\second} and 1 \unit{\radian\per\second}, respectively. \textbf{(i)} shows the trajectory taken by the trained policy to decelerate, stabilize, and avoid crashing into the walls. \textbf{(ii)} shows 20 \unit{\milli\second} event frames colored by polarity at different time instances. It is clearly visible that the event rates reduce over time. In this scenario, ego-motion of the quadrotor is the only source of intensity change in the camera frame, so as the vehicle decelerates and stabilizes, the event rates reduce significantly.
    }
\label{int}
\vspace{-1.5\baselineskip}
\end{figure}

Following \cite{das2025fastfeaturefield}, we train an RGB-based monocular depth prediction network (Depth Anything V2 \cite{depth_anything_v2}) that is modified to take $\text{F}^3$ as input. Training details are provided in the Supp. \cref{sec:depth_suppl}.
\cref{fig:neurosim_applications} (B) shows a qualitative example of the data (trajectory, events, RGB, and depth) generated from Neurosim and the depth predictions from the trained network.
The network is trained completely online. Each training step consumes a batch of data generated from Neurosim on-the-fly without any intermediate disk storage.
For context, the depth prediction model roughly consumes 8.5 M events per second, which would require $\sim$ 450 \unit{\giga\byte} of disk storage per hour if it was to be stored. Neurosim runs on a single Nvidia 4090 GPU roughly consuming 1.5 \unit{\giga\byte} of VRAM and 120 \unit{\watt} of power. The data generation rate is enough to saturate the training pipeline for an $\text{F}^3$-Depth Anything V2 ``small'' model running on the same machine on a second GPU.
This experiment demonstrates that Neurosim and Cortex can facilitate the training of foundation models for robot perception completely online on low-latency, high-throughput data.

\paragraph{Event-based Reinforcement Learning.}
Neurosim natively supports training event-based reinforcement learning (RL) policies completely online, enabling multiple concurrent simulation instances to produce high-throughput rollout data on a single GPU. Neurosim RL lowers the barrier of using multi-modal input, custom reward functions, domain randomization and collision checking, providing a modular interface to train both on and off-policy RL methods.
To demonstrate an on-policy training capability, we train a continuous-control policy via Proximal Policy Optimization (PPO) \cite{schulman2017proximalpolicyoptimizationalgorithms} for a ``stabilize-hover'' task. In this environment, a quadrotor is initialized with random linear and angular velocities. Relying solely on event-camera observations and state estimates, the vehicle must learn control actions to decelerate and stabilize to a hover without colliding with the scene. Both the structure and motion from event streams are critical to stabilize and not crash the quadrotor into obstables -- making the task particularly challenging and interesting for dynamic vision sensors.
The reward function is designed to penalize the magnitude of linear and angular velocities to encourage deceleration, while regularizing action smoothness for smoother control. Additionally, a flat-bonus survival reward is provided at each timestep to encourage the agent to avoid crashing and learn from longer trajectories.
Further algorithmic setup, reward specifications, and PPO hyperparameter details are provided in the Supp. \cref{sec:rl_suppl}. \cref{fig:neurosim_applications} (C) shows an example rollout of the trained policy. The policy is able to learn a trajectory that decelerates and stabilizes the quadrotor from a random initial velocities to hover without crashing in a cluttered indoor environment.

\section{Discussion}
\label{sec:discussion}

Neurosim and Cortex open up avenues to build and test new neuromorphic algorithms in simulation.
Multi-kilohertz event simulation and high-throughput multi-modal sensing with accurate calibration and scene ground-truth enable key aspects of training and evaluation in embodied autonomy. This supports (i) streaming data directly into learning and control pipelines without disk I/O; (ii) privileged supervision (e.g., metric depth, pose, and time-aligned cross-modal views) at high rates for online self-supervised learning and reinforcement learning that scales with environment diversity rather than dataset storage; (iii) closed-loop experiments at the limits of hardware performance for high-speed perception and control; and (iv) controlled studies of sensor timing, from synchronized streams to realistic clock skew and asynchronous sampling.

Several directions remain open to improvements. First, we plan to add hardware-in-the-loop experiments where an edge device (e.g., an embedded controller or perception pipeline on Jetson hardware) consumes simulated sensors and returns control at realistic latencies. This will expose timing, scheduling, and bandwidth constraints that are often abstracted away in typical evaluations. Second, we aim to support Unreal Engine 5 \cite{unrealengine} as an additional rendering backend to broaden asset compatibility and photorealism, while preserving the high-rate event generation pathway and maintaining determinism for reproducible learning.



{
    \small
    \bibliographystyle{ieeenat_fullname}
    \bibliography{bib/25_richeek_neurosim}
}

\clearpage
\setcounter{page}{1}
\maketitlesupplementary

\section{Online Training of Event-based Monocular Depth Estimation Model Details}
\label{sec:depth_suppl}

Following \cite{das2025fastfeaturefield}, we train an RGB-based monocular depth prediction network (Depth Anything V2 \cite{depth_anything_v2}) that is modified to take $\text{F}^3$ as input. We call this network $\rho: \reals^p \times \Om \to \reals \times \Om$, it takes $p$-channel $\text{F}^3(t,\cdot)$ as the input and predicts $\disp(t,\cdot) \in \reals\times \Om$ as the output. Since Neurosim provides dense ground-truth metric disparity $\disp^*(t, u)$ at any time $t$ and pixel $u$, we can train the network to predict disparity $\disp(t, u)$. Similar to the second stage of training in \cite{das2025fastfeaturefield}, we minimize a scale-invariant loss combined with a gradient-based regularizer to encourage sharp object boundaries:
\beq{
\aed{
    \ell_t(\rho) &=
       \frac{1}{|\Omega|}
    \sum_{u \in \Omega} \left(\log \frac{\disp(t, u)}{\disp^*(t,u)}\right)^2 \\
    &- \frac{1}{2 |\Omega|^2} \left(\sum_{u \in \Omega} \log \frac{\disp(t, u)}{\disp^*(t,u)}\right)^2
    + \lambda R (\tilde \disp, \tilde \disp^*),
    \label{eq:depth_objective}
}
}
where $\Omega$ is the set of pixels, and $\tilde \disp$ denotes the normalized disparity obtained by subtracting the median disparity from $\disp(t, u)$ and dividing by the average deviation from the median across $\Om$. The first two terms form the scale-invariant SiLog loss. The final term is the regularizer
\beq{
    R(\disp, \disp^*) = \f{1}{\lvert\Om\rvert} \sum_{\s} 2^{2 \s} \sum_u \lVert{\nabla_u \disp_\s(t, u) - \nabla_u \disp^*_\s(t, u)} \rVert_1,
    \label{eq:depth_regularizer}
}
which encourages accurate object boundaries by matching the gradient of the predictions with the gradient of the ground-truth disparity maps.
Averaging across spatial scales $\s$ (computed by down-sampling by a factor of two, without smoothing) encourages the network to use global context. We train this network for 400,000 steps with the same hyperparameters as those in \cite{das2025fastfeaturefield}.

\section{Event-Based Reinforcement Learning Setup Details}
\label{sec:rl_suppl}

In this section, we provide detailed information regarding the event-based reinforcement learning (RL) ``stabilize-hover'' task mentioned in \cref{sec:applications}.

\paragraph{Algorithm and Architecture.}
We train the policy using Proximal Policy Optimization (PPO) \cite{schulman2017proximalpolicyoptimizationalgorithms} -- implementation taken from the Stable-Baselines3 \cite{stable-baselines3} library. 
The policy network is trained on event representations using a CNN-based feature extractor and multirotor state estimates using a single layer fully-connected network. We utilize 8 concurrent environment instances simultaneously, created via subprocess vectorization to collect PPO rollout data.

\paragraph{Observation Space.}
The environment provides simulated events as $(n, 4)$ arrays representing the pixel locations, timestamps, and polarities. That is, $n$ variable number of events are collected at each time step, depending on the motion/intensity change due to ego-motion. We use a Time Surface \cite{timesurfaces2017} representation of the event stream as input to a CNN policy. For each event at time $t$, pixel $u \in \Om$, and polarity $p \in \{0, 1\}$, the time surface $S \in \R^{\Om \times 2}$ is updated based on the elapsed time $\Delta t$ since last such update:%
\begin{equation}
    S(x, y, p) \leftarrow S(x, y, p) \cdot \exp\left(-\frac{\Delta t}{\tau}\right) + 1.0,
\end{equation}%
where $\tau = 10 \unit{\milli\second}$ is the exponential decay time constant, $\Om$ is the spatial domain of the event camera ($480\times 640$ pixels in our experiments). $S$ is initialized to zero at the start of each episode. A Time Surface encodes the recency and frequency of events without requiring long temporal buffers. $\tau$ controls the memory of the time surface. Empirically, we find $10 \unit{\milli\second}$ of memory to be sufficient for the quadrotor speeds we operate at ($\sim 2.0 \unit{\meter\per\second}$). For more aggressive flights, noisier event streams, or complex tasks, learned representations such as $\text{F}^3$ \cite{das2025fastfeaturefield} can be used to prevent manually tuning $\tau$ and to extract more robust features from the raw event stream.

The policy also receives the multirotor state estimates (position, velocity, orientation, and angular velocity) as input. These are obtained from a simulated state estimator (RotorPy \cite{folk2023rotorpy}) that roughly mimics the noise characteristics of state estimation pipelines. The state estimates are provided at the same control frequency as the events (100 \unit{\hertz}), and are concatenated with the CNN features before being linearly projected onto the action space.

\paragraph{Action Space.}
The agent acts in a 4-dimensional continuous action space modeled as CTBR (collective thrust and body rates). The network predicts a normalized action in $[-1, 1]^4$, which is internally unnormalized by the environment to, (i) Collective thrust: Rescaled between the vehicle's minimum and maximum achievable thrusts, and (ii) Body rates: Roll, pitch, and yaw rates in the body frame, rescaled between the minimum and maximum achievable angular rates of the vehicle. These limits are set symmetrically (e.g., $\pm7$ \unit{\radian\per\second} for roll/pitch and $\pm5$ \unit{\radian\per\second} for yaw).

\paragraph{Reward Design.}
The training objective is to guide the multirotor to a stable hover state while avoiding collisions with the scene. The reward function $r$ utilizes potential-based shaping alongside Gaussian kernels to encourage zero-velocity states. We define the reward at time $t$ as a sum of five components,
\beq{
    r = r_{\text{velocity}} + r_{\text{angular}} + r_{\text{action}} + r_{\text{survival}},
}
where:
\begin{itemize}
    \item $r_{\text{velocity}} = - w_{\text{velocity}} \norm{v}_2$ penalizes the magnitude of the linear velocity $v$ to encourage the agent to slow down.
    \item $r_{\text{angular}} = - w_{\text{angular}} \norm{\w}_2$ similarly penalizes the magnitude of the angular velocity $\w$ to encourage the agent to stabilize its orientation.
    \item $r_{\text{action}} = -w_{\text{action}} \norm{a_t - a_{t-1}}_2$ penalizes jerk, encouraging smoother actions and preventing erratic control.
    \item $r_{\text{survival}} = w_{\text{survival}}$ is a constant positive reward at each time step to encourage the agent to survive longer and learn from more data. This is crucial in the early stages of training when the agent's policy is still random and prone to crashing immediately.
\end{itemize}

The reward weights are assigned based on the relative importance of each reward component in guiding the policy towards stable hovering. We choose  $w_{\text{velocity}}=2.0$, $w_{\text{angular}}=0.05$, $w_{\text{action}}=0.01$ and $w_{\text{survival}}=0.5$. Collisions result in immediate episode termination and a substantial negative penalty of $-500$, roughly equivalent to the cumulative reward of surviving for $10 \unit{\second}$ without crashing. This encourages the agent to prioritize collision avoidance while learning to stabilize. Episodes only terminate upon collision or after a maximum episode length of $15 \unit{\second}$.

\paragraph{Implementation and Hyperparameters.}
We utilize 8 independent concurrent simulation instances running as subprocesses on a single NVIDIA RTX 4090 GPU to collect real-time rollout data. The PPO policy is trained for a total of 3 M environment steps.

At each iteration, we collect rollouts using 512 steps per environment and optimize the policy over 8 epochs with a minibatch size of 128. We configure PPO with a learning rate of $3 \times 10^{-4}$ and apply an objective clipping range of 0.2. The Generalized Advantage Estimation (GAE) procedure utilizes scaling constants $\gamma=0.99$ and $\lambda=0.95$. Other algorithmic parameters include an entropy coefficient of $0.003$ to encourage moderate exploration, a value function coefficient of $0.5$, and a maximum gradient norm of $0.5$.

\end{document}